\documentclass{article}
\usepackage{arxiv}
\usepackage{amsfonts,booktabs,multirow,lineno,hyperref,xcolor,amsmath}
\usepackage{blkarray, bigstrut,pifont}
\usepackage{booktabs}
\usepackage{amsthm,subfigure}
\usepackage{hyperref,booktabs}
\usepackage{algorithmic}
\usepackage{algorithm}

\usepackage[utf8]{inputenc} 
\usepackage[T1]{fontenc}    
\usepackage{hyperref}       
\usepackage{url}            
\usepackage{booktabs}       
\usepackage{amsfonts}       
\usepackage{nicefrac}       
\usepackage{microtype}      
\usepackage{lipsum}
\usepackage{graphicx}
\graphicspath{ {./images/} }

\usepackage{amsmath,amssymb,bm}
\usepackage{adjustbox}
\usepackage{caption}
\usepackage{subfigure}
\usepackage{multirow}
\usepackage[english]{babel}
\usepackage{amsthm}

\title{Efficient Hybrid Oversampling and Intelligent Undersampling for Imbalanced Big Data Classification}

\author{
Carla Vairetti \\
Universidad de los Andes, Chile \\
Facultad de Ingenier\'{i}a y Ciencias Aplicadas\\
\& Instituto Sistemas Complejos de Ingenier\'\i a (ISCI), Chile.\\
\texttt{cvairetti@uandes.cl}
\And
 Jos\'e Luis Assadi \\
 Universidad de los Andes, Chile\\
Facultad de Ingenier\'{i}a y Ciencias Aplicadas,\\
\texttt{jlassadi@miuandes.cl}
\And
Sebasti\'an Maldonado\\
Department of Management Control and Information Systems \\
School of Economics and Business\\
University of Chile, Chile\\
\& Instituto Sistemas Complejos de Ingenier\'\i a (ISCI), Chile.\\
\texttt{sebastianm@fen.uchile.cl}
}

\begin{document}
\maketitle

\begin{abstract}
Imbalanced classification is a well-known challenge faced by many real-world applications. This issue occurs when the distribution of the target variable is skewed, leading to a prediction bias toward the majority class. With the arrival of the Big Data era, there is a pressing need for efficient solutions to solve this problem. In this work, we present a novel resampling method called SMOTENN that combines intelligent undersampling and oversampling using a MapReduce framework. Both procedures are performed on the same pass over the data, conferring efficiency to the technique. The SMOTENN method is complemented with an efficient implementation of the neighborhoods related to the minority samples. Our experimental results show the virtues of this approach, outperforming alternative resampling techniques for small- and medium-sized datasets while achieving positive results on large datasets with reduced running times. 
\footnote{This is a preprint of a work under submission and thus subject to change. Changes resulting from the publishing process, such as editing, corrections,structural formatting, and other quality control mechanisms may not be reflected in this version of the document.}
\end{abstract}

\section{Introduction}

Binary classification is a relevant topic in machine learning due to its wide range of application areas, and the design of new algorithms has resulted in a fruitful research area \cite{Fernandez18-Book}. The performance of the classifiers, however, depends not only on the selected algorithm but also on the quality of the input data \cite{GARCIAGIL2019135,Triguero19-SmartD}. In this sense, the class imbalance problem is an important issue that arises in many tasks \cite{Vanhoeyveld2018,ZHU20221397}. This problem occurs when one of the two classes is underrepresented with respect to the other. 

Imbalanced Big Data classification has been acknowledged as a relevant open challenge in machine learning \cite{krawczyk2016learning}. The most popular strategies for dealing with the class-imbalance issue, such as random undersampling (RUS) and the synthetic minority oversampling technique (SMOTE), have been adapted for large datasets \cite{juez2021approx}. These techniques preprocess the data, either by downsizing the majority class (undersampling) or by creating synthetic examples from the minority class (SMOTE and other oversampling variants). 

Dealing with large-scale datasets effectively is of utmost importance in predictive analytics and, in particular, in binary classification tasks. This is because of the varied and complex nature of large datasets and the recent need to handle hundreds or even millions of variables from different sources to explain a given phenomenon \cite{fernandez2017insight}. Predictive models for classification tasks are usually more time-consuming than other data analysis tasks, such as data storage and dashboarding, and traditional methods become intractable in terms of running times when facing large datasets. To address this challenge, distributed machine learning approaches have been developed. This ``divide and conquer'' strategy is designed to scale to larger input data sizes via multinode algorithms and systems \cite{KADKHODAEI2021115369}. To this end, the MapReduce framework has become popular as a Big Data tool for machine learning, and it has been used for dealing with the class-imbalance problem \cite{fernandez2017insight}.

There is an important gap in the imbalanced big data classification literature. This is because no hybrid undersampling-oversampling methods have been proposed, to the best of our knowledge, and only a few intelligent undersampling techniques have been reported in the literature. This research is designed to fill this gap. On the one hand, the combination of two or more resampling methods has shown the best performance in relation to using a single strategy \cite{krawczyk2016learning,xu2020hybrid}. On the other hand, there are several undersampling strategies that remove examples from the majority class in an ``intelligent'' manner, i.e., by exploiting the structure of the data instead of simply discarding examples randomly. For example, the edited nearest neighbor (ENN) approach \cite{Wilson1972} ``cleans'' a neighborhood of samples from the minority class by eliminating majority-class examples within its boundaries \cite{batista2004study}.

The main issue with hybrid methods is that combining strategies can be time-consuming when performed independently. To overcome this issue, we propose a novel hybrid approach that combines SMOTE and ENN in a single pass over the data. These two methods have in common that a neighborhood is defined; however, the computation of distances can be the most time-consuming task for resampling techniques in Big Data environments \cite{juez2021approx}. We propose a method called SMOTENN that defines a single neighborhood for each example in the minority class, in which both the creation of synthetic examples and the elimination of majority class samples are performed. This method can be combined further with RUS to provide suitable preprocessing for binary classification tasks. 

In summary, the main contributions of this study are as follows:
\begin{itemize}
\item We present a novel hybrid undersampling-oversampling method for imbalanced classification. To the best of our knowledge, this is the first hybrid undersampling-oversampling approach for large-scale machine learning, but it is also suitable for datasets of regular size.
\item We conducted a comprehensive experimental study on 35 datasets of different sizes, analyzing the performance of several resampling methods. Our results show that SMOTENN is able to outperform other resampling methods for small- and medium-sized datasets. For large datasets, our method has a positive performance on average compared to other imbalanced Big Data techniques, performing better than RUS and the combination of RUS and SMOTE. 
\item We empirically discuss and evaluate the value of creating new examples from the minority class in a Big Data context. Most studies on imbalanced Big Data classification focus either on undersampling \cite{triguero2016evolutionary} or oversampling \cite{juez2021approx}, without discussing the differences between these two approaches. 
\end{itemize}

The remainder of this study is structured as follows: Section \ref{sec:prior} discusses prior work on class-imbalanced classification, providing the preliminaries for the SMOTENN. Next, the proposed hybrid undersampling-oversampling method is formalized in Section \ref{sec:prop}. The experimental results obtained for small, medium, and large benchmark datasets are presented in Section \ref{sec:exp}. Finally, Section \ref{sec:conc} provides the main conclusions of this paper.

\section{Prior work}\label{sec:prior}

The two topics relevant to this proposal are discussed next. First, class-imbalanced classification methods are presented, with emphasis on data resampling techniques. The second topic is imbalanced Big Data classification, in which a relevant framework for distributed machine learning and other solutions for handling large datasets are discussed. This section concludes with an analysis of the state of the art on imbalanced Big Data classification.

\subsection{Resampling techniques for class-imbalanced classification}

There are two main approaches for dealing with the class imbalance issue: data resampling and algorithmic-level solutions \cite{Fernandez18-Book}. Data resampling consists of balancing the training set as a preprocessing step, either by downsizing the majority class (undersampling) or by creating synthetic examples from the minority class (oversampling) \cite{kamal2016mapreduce,YAN2022116213}. There are very simple techniques for this task, such as RUS but also intelligent approaches that take into account important pitfalls that occur when facing highly imbalanced datasets \cite{YAN2022116213}. 

Algorithmic-level solutions are adaptations made to classifiers to learn from imbalanced datasets without data resampling \cite{Fernandez18-Book}. There are different approaches for achieving this goal, such as cost-sensitive methods and one-class learning approaches \cite{Fernandez18-Book}. Although there are a plethora of algorithmic-level solutions for dealing with the class-imbalance problem, these strategies fall outside the scope of this paper. Finally, hybrid strategies combine two or more techniques to find an adequate compromise between the two worlds \cite{krawczyk2016learning}. 

\subsubsection*{Undersampling}

Random undersampling is the simplest solution for the class-imbalance problem and consists of randomly excluding examples of the majority class from the training set. This technique can be performed without significant computational effort, making it suitable in distributed environments. Furthermore, RUS leads to classifiers that can be trained faster than when using the original datasets \cite{fernandez2017insight,Leevy2018}. However, this technique has some limitations. On the one hand, this implies ignoring a vast amount of training data and discarding potentially useful observations. On the other hand, this loss of information may distort the underlying distribution of the majority sampling \cite{fernandez2017insight,Leevy2018}.

To overcome the shortcomings associated with RUS, some intelligent undersampling techniques have been proposed. As indicated in the previous section, the ENN method removes majority-class samples from a neighborhood as a data-cleansing step \cite{batista2004study}. This is arguably the best-known intelligent undersampling technique. The algorithm works as follows:

\begin{enumerate}
\item Define the size of the neighborhood $K$. The default number is $K=3$
\item Find the $k$-nearest neighbors ($k$-NN) of a given sample and identify the majority class from the neighborhood.
\item If the class of the sample does not coincide with the majority class of the neighborhood, then both the sample and the neighborhood are deleted.
\item Repeat steps 2 and 3 until a predefined imbalance ratio (IR) is achieved. 
\end{enumerate}

The ENN algorithm can be iterated multiple times, shrinking the majority class with each pass. The original method by \cite{Wilson1972} does not specify a clear stopping criteria. A possible approach is to continue until reaching a predetermined imbalance ratio (IR), defined as the size of the majority class divided by the size of the minority class. For instance, one might iterate until the majority and minority classes each comprise 50\% of the samples (IR=1).

Several intelligent undersampling strategies have been proposed in recent years. For example, evolutionary undersampling (EUS) involves intelligent resampling methods that utilize evolutionary computation to determine which samples should be discarded. Evolutionary computation techniques emulate the behaviors of living organisms to solve tasks \cite{triguero2015evolutionary}. For instance, Genetic Algorithms (GAs), inspired by natural selection, are commonly applied in EUS. GAs can efficiently tackle optimization problems, like those in clustering-based resampling. Clustering models typically present an optimization problem where the goal is to minimize the within-cluster variance, a measure of dispersion for samples within that cluster. In \cite{Kim2016}, a clustering model is trained using a GA, where samples distant from the cluster centroid are viewed as noise. These noisy examples from the majority class are subsequently removed, leading to an efficient undersampling method.

\subsubsection*{Oversampling}

The simplest oversampling strategy is random oversampling, in which examples of the minority class are replicated. This is equivalent to introducing a higher weight to these samples in the case of misclassification \cite{Fernandez18-Book}. This strategy, however, is prone to overfitting \cite{Nekooeimehr2016}.

From the myriad of different oversampling strategies, SMOTE is arguably the best-known strategy \cite{Fernandez18-Book}. It creates new instances that result from interpolating among neighboring minority class samples \cite{Cha2}. Formally, given a sample $\textbf{x}_i$ from the minority class and one randomly-chosen example from its neighborhood $\textbf{x}_i^p$, with $p=1,\ldots,K$, a new synthetic sample $\textbf{x}_i^{*p}$ is obtained using the following expression:

\begin{equation} \label{SMOTE}
\textbf{x}_i^{*p}:=\textbf{x}_i+u\left(\textbf{x}_i^p-\textbf{x}_i \right),
\end{equation}
where $u$ is a randomly generated number between 0 and 1. The iterative process behind SMOTE ends when the desired oversampling amount $N<K$ is reached. For example, if the goal is to triplicate the minority class ($N=2$), a neighborhood of size $K=5$ can be considered, and two examples can be randomly chosen from it, one at the time. With 1000 examples in the minority class, for instance, this results in a total 3000 examples for this class, comprising the original ones and an additional 2000 synthetically-generated samples.

SMOTE has the advantage of being fast to compute and successful at providing balanced and accurate classification performance, reducing the risk of overfitting. Due to its popularity, different SMOTE extensions have been proposed to improve some of its shortcomings. ADASYN \cite{He2008}, for example, selects the amount of oversampling for each minority example dynamically by estimating its intrinsic difficulty, which is based on the ratio of examples belonging to the majority class in the neighborhood. 

Some SMOTE variants focus on the selection of instances that are closer to the boundary areas. Borderline-SMOTE \cite{Han2005} is one example that is based on the premise that the examples far from the borderline may not provide a strong contribution to the classification ability of the model. Density-based SMOTE \cite{bun11} also operates in an overlapping region but considers a density-based approach of clustering called DBSCAN, generating synthetic instances by computing the shortest path from each minority sample to a pseudocentroid of a minority-class cluster. 

Recent oversampling approaches involve the use of deep learning models such as generative adversarial networks (GANs) \cite{ZHU20221397} to create synthetic examples. This strategy confers the flexibility to create more sophisticated synthetic examples, such as images. However, training an additional machine learning model for preprocessing can be time-consuming in Big Data settings.

As indicated earlier in this paper, undersampling and oversampling approaches can be combined to alleviate the need to either create too many synthetic samples or remove an important proportion of majority-class examples. Arguably, the most common combination is RUS with SMOTE. ENN and SMOTE were proposed in combination in \cite{batista2004study}. This combination of approaches has shown great success in recent studies \cite{kandula2021prescriptive,wang2021improving,xu2020hybrid}. However, the implementation designed in \cite{batista2004study} obtains different neighborhoods for the undersampling and oversampling steps, resulting in two almost independent processes. This study aims to overcome this issue to scale up this method to Big Data environments. 

\subsection{Imbalanced Big Data classification}

Gaining critical insights by querying and analyzing massive amounts of data has become a challenge to standard data science approaches. In this sense, big data can be defined as data that exceed the processing capacity of conventional systems, and therefore, new techniques designed to process the information are needed \cite{delRio2014,KADKHODAEI2021115369}. 

The MapReduce execution environment \cite{dean2008mapreduce} is the most common framework that has been traditionally used in academic research to provide robust and scalable solutions for big data problems \cite{white2012hadoop}. This framework has been used in imbalanced Big Data classification, and some resampling techniques have been adapted to face large-scale problems. Resampling approaches such as RUS, random oversampling, and SMOTE have been embedded in a MapReduce framework \cite{fernandez2017insight}.

SMOTE was further improved in terms of scalability in \cite{juez2021approx}. In this work, the authors propose Approx-SMOTE, which considers an approximated version of the $k$-nearest neighbor algorithm to define the neighborhood before performing data interpolation. SMOTE was also extended to Big Data classification under a multiclass setting in \cite{SLEEMANIV2021106598}. Finally, evolutionary undersampling was also extended using MapReduce \cite{triguero2016evolutionary}. For these approaches, the random forest has been shown to be very effective as a baseline classifier.

Some advances have been developed for dealing with imbalanced Big Data classification tasks via algorithmic and cost-sensitive approaches. For example, the random forest was extended to cost-sensitive imbalanced Big Data classification in \cite{delRio2014}. Extreme learning machine (ELM) is another method that has been adapted as a cost-sensitive learning method for distributed environments \cite{ALABA201970}. Finally, fuzzy rule-based classifiers have been adapted for imbalanced Big Data classification thanks to the use of the MapReduce framework \cite{lopez2015cost}.

In summary, the following gaps in the literature were identified:
\begin{itemize}
 \item Although there are many existing intelligent oversampling techniques, few studies are devoted to intelligent undersampling. There are plenty of opportunities for improving classification performance using novel intelligent undersampling techniques. 
 \item Although undersampling can be beneficial for reducing the complexity of classifiers in Big Data classification environments, there are few studies using this approach compared to oversampling. 
 \item Hybrid undersampling/oversampling approaches represent an interesting line of research and have been shown to be effective when dealing with the class-imbalance problem. However, most hybrid approaches perform both strategies independently. It would be interesting to combine these strategies in an intelligent manner in imbalanced big data classification environments.
\end{itemize}

\section{Proposed SMOTENN method for hybrid data resampling}\label{sec:prop}

The main goal of the proposed technique is to perform data resampling for imbalanced classification efficiently, considering 1) a distributed framework for data loading and processing, 2) a fast and scalable distance metric to define a neighborhood of samples, and 3) a novel algorithm able to perform both SMOTE-like oversampling and intelligent undersampling in a single pass over the data. To achieve this, a neighborhood of both majority and minority samples is defined for each example from the minority class. We consider ENN as the undersampling method, which reduces the noise in the neighborhood before performing oversampling via interpolation. 

This section is structured according to the three aspects of the proposed method described in the last paragraph. Section \ref{ss_Mapreduce} discusses the MapReduce framework considered for efficient data loading and processing. Next, the fuzzy $k$ nearest neighbors (NN) algorithm proposed in \cite{Maillo2020} and its adaptation to hybrid data resampling are presented in Section \ref{ss_AproxkNN}. Finally, the SMOTENN algorithm is formalized in Section \ref{ss_SMOTENN}.

\subsection{The MapReduce Framework}\label{ss_Mapreduce}

 Traditional systems use a centralized server for storing and retrieving data. However, a large amount of data cannot be accommodated in standard database servers when facing Big Data. Furthermore, centralized systems tend to generate a bottleneck when processing multiple files simultaneously. The MapReduce algorithm solves this issue by dividing the tasks into small parts and processing each part independently by assigning them to different systems. After the parts are processed and analyzed, the output of each computer is collected in a single location, and then an output dataset is prepared for the given problem \cite{fernandez2014big, srivas2016map}.

\begin{figure}[ht!]
 \centering
 \caption{The MapReduce framework.}
 \includegraphics[width=0.99\textwidth]{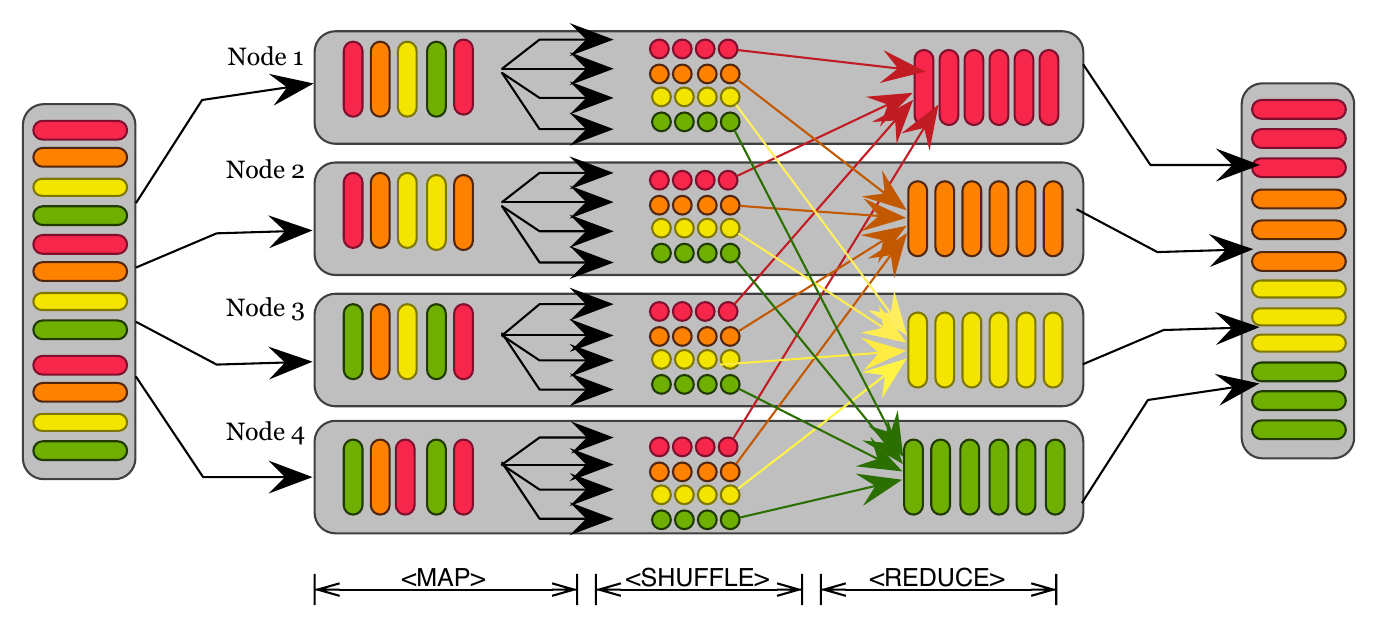}
 \vspace{-0.5cm}
 \label{mapReduce}
 \end{figure}

The basic unit information for structured or unstructured data used by MapReduce is a key-value pair \cite{fernandez2017insight}. The MapReduce model has three main steps: map, shuffle, and reduce. The ``map'' step takes as input a function and a sequence of values and applies the function to each value in the sequence, filtering and sorting it according to predefined parameters. The mapped data are output to a temporary storage with a corresponding set of keys, indicating how the data should be redistributed. The ``shuffle'' step consists of redistributing the mapped data to other nodes in the system so that each node contains groups of key-similar data. Finally, the ``reduce'' step takes as input a sequence of elements and combines all the elements using a binary operation. These steps are illustrated in Figure \ref{mapReduce}.

 In-memory Spark and Apache Spark have emerged as popular choices for implementing large-scale machine learning applications in Big Data environments. These tools are included in the Hadoop ecosystem \cite{spark2016apache}. Spark is particularly efficient thanks to its approach for representing data in memory, called the resilient distributed dataset (RDD). This data structure technique is a clever way of guaranteeing fault tolerance that minimizes network I/O \cite{spark2016apache}.

Created as a Spark subproject, the MLlib library includes common learning algorithms and statistical utilities for Big Data and machine learning \cite{meng2016mllib}. Its main functionalities include classification, regression, clustering, collaborative filtering, optimization, and dimensionality reduction (mostly feature extraction). Thanks to Spark's capacity for in-memory computation that speeds up iterative processes, algorithms developed for this platform have become pervasive in the industry. Both Spark and MLlib are considered in the proposed method for the implementation of the data resampling techniques and the machine learning strategies when facing imbalanced classification. 

\subsection{An efficient $k$-NN search for SMOTENN}\label{ss_AproxkNN}

A key component of the proposed algorithm is the use of a fast and scalable distance metric. This measure was used for fuzzy $k$-NN classification in \cite{Maillo2020} for classification and was adapted to data resampling in \cite{juez2021approx}. The approximated distance function considers the hybrid spill tree approach, which splits the feature space so that the examples that fall in the same leaf can be considered neighbors. This approach is hybrid in the sense that it combines two strategies: the metric tree and the spill tree.

A metric tree is a structure for data indexing designed for an efficient $k$-NN search. It is a binary tree that organizes the data points hierarchically in such a way that the children of a node are two disjoint sets taken as far as possible. For a given data point, the method keeps a candidate for the nearest neighbor and the distance between the two samples $d$. In case the distance to a branch is greater than $d$, the algorithm prunes this branch and continues the search \cite{Maillo2020}. 

For a given node $\nu$, let $\mathcal{N}(\nu)$ be all the samples at this node, and let $left(\nu)$ and $right(\nu)$ be the left and right children, respectively. A random point from $\mathcal{N}(\nu)$ is selected, and its furthest point $p_l$ is identified. Let $p_r$ be the furthest point from $p_l$. Next, all samples between these two points are projected, and the median is identified. All points on the left of this median go to $left(\nu)$, while the remaining points go to $right(\nu)$, defining a boundary $L$.

The metric tree algorithm is very efficient in finding the nearest neighbors quickly; however, it spends a large amount of time verifying whether this NN is the right one. To speed up the process, spill trees allow overlapped sets in the binary trees to find an approximate answer in the search for the NN. Therefore, the children of a node can ``spill over'' onto each other and share examples \cite{Maillo2020}.

For a given node $\nu$, let $\tau$ be the area that both children share. Then, the method defines two boundaries, $L_l=L-\tau$ and $L_r=L+\tau$. All points on the left of $L_l$ go to $left(\nu)$, while those on the right of $L_r$ go to $right(\nu)$.

Although this approximate approach is able to solve the issue of verifying the NN that metric trees face, the depth of the spill tree varies considerably depending on the value of $\tau$. To solve both issues, hybrid spill trees combine these strategies and introduce a balance factor $0 \leq \rho < 1$.

For a given node $\nu$, if either of its two children contains more than $\rho |\mathcal{N}(\nu)|$, then the metric tree procedure is followed, and nonoverlapping nodes are defined. Otherwise, the spill tree procedure is considered, leading to overlapped children. The backtracking strategy that verifies the NN is only performed for nonoverlapping nodes \cite{Maillo2020}. 

\subsection{The SMOTENN algorithm}\label{ss_SMOTENN}

Undersampling can be valuable for reducing the overall complexity of the learning process. However, several studies have shown that hybrid undersampling/oversampling approaches can be useful for improving the predictive performance of classifiers. In particular, information from the majority class can be useful for improving both undersampling and oversampling strategies, such as in ENN, the Borderline SMOTE, and SafeLevel SMOTE methods discussed in the previous section.

Having defined the approximated distance metric and the MapReduce framework, we now describe the novel data resampling algorithm for imbalanced Big Data classification. This strategy combines RUS, ENN for intelligent undersampling, and a variation of the SMOTE oversampling. The proposed SMOTENN method is formalized in Algorithm \ref{algo1}.

\begin{algorithm}[!ht]
\caption{SMOTENN method for imbalanced Big Data classification}
\textbf{Input:} Minority and majority class sample sets $\mathcal{S}_+$ and $\mathcal{S}_-$, respectively; Amount of oversampling $N$; Number of nearest neighbors $K$; Proportion of undersampling $p$; SMOTE parameter $\upsilon$.\\
\textbf{Output:} Resampled set of elements $\mathcal{S}^*$.
\begin{enumerate}
\item $\mathcal{S}_-^* \ \leftarrow$ Select a random sample of $p$\% of $\mathcal{S}_-$.
\item $\mathcal{S}^* \leftarrow \mathcal{S}_-^* \cup \mathcal{S}_+$. 
\item $\mathcal{L}\leftarrow \emptyset$ Initialize the set of removed examples from the majority class for the ENN step.
\item \textbf{for} $ \ i \ \in \mathcal{S}_+$ 
\item $\ \ \ \mathcal{T}_i \leftarrow$ Set of the $K$ nearest neighbors of $i$ in $\mathcal{S}_-^* \cup \mathcal{S}_+ \setminus \{i\}$.
\item $\ \ \ $\textbf{if} the majority of the neighbors of $i$ are from $\mathcal{S}_+$ \textbf{then}
\item $\ \ \ \ \ \ \mathcal{L} \leftarrow \ \mathcal{L} \cup (\mathcal{T}_i \cap \mathcal{S}_-^*)$.
\item $\ \ \ \ \ \ \mathcal{T}_i \leftarrow \ \mathcal{T}_i \cap \mathcal{S}_+$. 
\item $\ \ \ \ \ \ $\textbf{for} $ \ k \leftarrow 1$ \textbf{ to } $N$ 
\item $\ \ \ \ \ \ \ \ \ \textbf{x}_{k} \leftarrow$ Select a random sample from $\mathcal{T}_i$.
\item $\ \ \ \ \ \ \ \ \ \textbf{x}^{*}_{k} \leftarrow \textbf{x}_{i} + \upsilon (\textbf{x}_{k}-\textbf{x}_{i})$. 
\item $\ \ \ \ \ \ \ \ \ \mathcal{S}^* \leftarrow (\mathcal{S}^*, \textbf{x}^{*}_{k})$.
\item $\ \ \ \ \ \ \ \ \ \mathcal{T}_i \leftarrow \mathcal{T}_i \setminus \{\textbf{x}_{k}\}$. 
\item $\ \ \ \ \ \ $\textbf{end for}
\item $\ \ \ $\textbf{end if}
\item \textbf{end for}
\item $\mathcal{S}^* \leftarrow \mathcal{S}^* \setminus \mathcal{L}$. 
\end{enumerate}
\label{algo1}
\end{algorithm}

Next, a step-by-step description of the SMOTENN algorithm is presented:

\begin{itemize}
\item The inputs of the algorithm are the sample sets $\mathcal{S}_+$ and $\mathcal{S}_-$, the amount of oversampling $N$ and SMOTE parameter $\upsilon$ associated with the SMOTE step, the number of neighbors $K$, and the proportion of undersampling $p$. Parameters $N$, $K$, and $\upsilon$ are usually fixed values in traditional SMOTE: $N=\{1,3\}$ (duplication or quadruplication of the minority class), $K=5$, and $\upsilon$ is a random number between 0 and 1. We consider $N$ and $\upsilon$ as fixed parameters as well, but the number of neighbors should be considered as a tuning parameter before we define a default configuration. In contrast to SMOTE, SMOTENN defines a neighborhood that considers minority and majority class samples, and therefore, a neighborhood of a size greater than 5 may be needed. We also suggest performing RUS before the main loop that combines SMOTE and ENN in the case of extreme imbalances. The amount of RUS undersampling $p$ is also considered a tuning parameter. 
\item Step 1 of SMOTENN is the RUS step applied on $\mathcal{S}_-$ (the majority class set). Next, a new set $\mathcal{L}$ that includes all removed samples is initialized as an empty set. 
\item Steps 4-16 represent the main loop of the algorithm, which can be divided into the ENN step (Steps 7-8) and the SMOTE step (Steps 9-14). For each sample in $\mathcal{S}_+$, its $K$ nearest neighbors are identified using the approximated distance defined in Section \ref{ss_AproxkNN}. In contrast to performing ENN and SMOTE independently, a neighborhood is computed only once for each minority class sample. This is done in Step 5.
\item The condition in Step 6 suggests that ENN and SMOTE are performed only when most of the neighbors of $i$ belong to the minority class ($\mathcal{S}_+$). Here, the majority class samples are few in the neighborhood and can be viewed as noise. We first remove these majority class samples (ENN step) and then generate synthetic minority class examples through interpolation (SMOTE step). If the neighborhood contains more samples from the majority class than from the minority class, no resampling is performed.
\item Steps 7-8 formalize the ENN step: all the examples in the neighborhood $\mathcal{T}_i$ that belong to the majority class ($\mathcal{T}_i \cap \mathcal{S}_-^*$) are eliminated from the neighborhood and included in the list of removed samples $\mathcal{L}$. 
\item Steps 9-14 formalize the SMOTE step: a random sample $\textbf{x}_{k}$ from $\mathcal{T}_i$ is selected (Step 10) and used in combination with $\textbf{x}_{i}$ to create a new example $\textbf{x}^{*}_{k}$ via linear interpolation (Step 11). Note that $\mathcal{T}_i$ contains only minority class examples after Step 8. The synthetic example is appended to the final resampled set $\mathcal{S}^*$, and $\textbf{x}_{k}$ is removed from the neighborhood to avoid repetition. This process is repeated $N$ times.
\item After all the neighborhoods associated with the minority class samples are explored, the removed samples $\mathcal{L}$ are excluded from $\mathcal{S}^*$, leading to the final sample set. 
\end{itemize}

The process described in Algorithm \ref{algo1} can be repeated using different blocks of data (batches), constructing the final dataset for training in an incremental fashion. This is suitable with the ``divide-and-conquer'' framework proposed for imbalanced Big Data classification using MapReduce.

\section{Experimental results}\label{sec:exp}

The proposed resampling method was applied to 35 benchmark datasets for class-imbalance classification. This section is structured as follows: Section \ref{ss_dataset} provides the experimental setting, while a summary of the results achieved with SMOTENN and the alternative methods is presented in Section \ref{ss_summary1} for small and medium-size datasets and in Section \ref{ss_summary2} for large datasets. The experimental section concludes with a sensitivity analysis for the main SMOTENN parameters and a discussion on the running times. This analysis is reported in Section \ref{ss_sensitivity}. 

\subsection{Experimental setting}\label{ss_dataset}

Regarding the datasets considered in this study, we consider three different types based on the sample size $m$: small datasets ($sds$, $m<1000$), medium-size datasets ($mds$, $1000 \leq m<100,000$), and large datasets ($lds$, $m \geq 100,000$). The small datasets are available in the UCI (\url{http://archive.ics.uci.edu/}) and KEEL (\url{http://sci2s.ugr.es/keel}) data repositories.

Table \ref{tab1}, Table \ref{tab2}, and Table \ref{tab3} summarize the relevant information for each dataset for the small, medium-size, and large datasets, respectively. Each table includes the sample size $m$, the number of variables $n$, the percentage of examples in each class (min., maj.), and the imbalance ratio (IR). 

\begin{table}[ht!]
\centering
\caption{Descriptive information for the small datasets ($sds$). This information includes the sample size $m$, the number of variables $n$, the proportion of examples from the minority and majority class \%class(min,maj), and the imbalance ratio $IR$.} \label{tab1}
\resizebox{.99\textwidth}{!}{
\begin{tabular}{llcccc}
\toprule
 code & Dataset & $m$ & $n$ & \%class(min,maj) & $IR$ \\
\midrule
 sds1 & glass-0-4 vs 5 & 92 & 9 & (9.8. 90.2) & 9.22 \\
 sds2 & glass-0-6 vs 5 & 108 & 9 & (91.7. 8.3) & 11 \\
 sds3 & glass-0-1-6 vs 2 & 192 & 9 & (8.9. 91.1) & 10.3 \\
 sds4 & ecoli-0-3-4 vs 5 & 200 & 7 & (10.0. 90.0) & 9 \\
 sds5 & ecoli-0-2-3-4 vs 5 & 202 & 7 & (9.9. 90.1) & 9.1 \\
 sds6 & ecoli-0-4-6 vs 5 & 203 & 6 & (9.9. 90.1) & 9.15 \\
 sds7 & ecoli-0-3-4-6 vs 5 & 205 & 7 & (9.8. 90.2) & 9.25 \\
 sds8 & glass4 & 214 & 9 & (6.1. 93.9) & 15.5 \\
 sds9 & ecoli-0-6-7 vs 5 & 220 & 6 & (9.1. 90.9) & 10 \\
 sds10 & ecoli-0-6-7 vs 3-5 & 222 & 7 & (9.9. 90.1) & 9.09 \\
 sds11 & ecoli-0-2-6-7 vs 3-5 & 224 & 7 & (9.8. 90.2) & 9.18 \\
 sds12 & ecoli-0-1 vs 5 & 240 & 6 & (8.3. 91.7) & 11 \\
 sds13 & ecoli-0-3-4-7 vs 5-6 & 257 & 7 & (9.7. 90.3) & 9.28 \\
 sds14 & ecoli-0-1-4-6 vs 5 & 280 & 6 & (7.1. 92.9) & 13 \\
 sds15 & ecoli-0-1-4-7 vs 5-6 & 332 & 6 & (7.5. 92.5) & 12.3 \\
 sds16 & ecoli & 336 & 7 & (10.4. 89.0) & 8.6 \\
 sds17 & ecoli4 & 336 & 7 & (6.7. 93.3) & 13.8 \\
 sds18 & yeast-1 vs 7 & 459 & 7 & (6.7. 93.3) & 13.9 \\
 sds19 & page-blocks-1-3 vs 4 & 472 & 10 & (5.9.94.1) & 15.9 \\
 sds20 & yeast-2 vs 4 & 514 & 8 & (9.9. 90.1) & 9.08 \\
 sds21 & yeast-0-5-6-7-9 vs 4 & 528 & 8 & (9.7.90.3) & 9.35 \\
 sds22 & yeast-1-4-5-8 vs 7 & 693 & 8 & (4.3.95.7) & 22.1 \\
 \bottomrule
\end{tabular}}
\end{table}

\begin{table}[ht!]
\centering
\caption{Descriptive information for the medium-size datasets ($mds$). This information includes the sample size $m$, the number of variables $p$, the proportion of examples from the minority and majority class \%class(min,maj), and the imbalance ratio $IR$.} \label{tab2}
\resizebox{.99\textwidth}{!}{
\begin{tabular}{llcccc}
\toprule
 code & Dataset & $m$ & $n$ & \%class(min,maj) & $IR$ \\
\midrule
 mds1 & yeast-0-2-5-7-9 vs 3-6-8 & 1004 & 8 & (9.9. 90.1) & 9.14 \\
 mds2 & solar & 1389 & 10 & (4.9.95.1) & 19.4 \\
 mds3 & yeast3 & 1484 & 8 & (11.0. 89.0) & 8.1 \\
 mds4 & yeast4 & 1484 & 8 & (3.4.96.6) & 28.1 \\
 mds5 & yeast5 & 1484 & 8 & (3.0. 97.0) & 32.78 \\
 mds6 & shuttle-c0-vs-c4 & 1829 & 9 & (6.7. 93.3) & 13.87 \\
 mds7 & image1 & 2310 & 19 & (14.3.85.7) & 6 \\
 mds8 & abalone7 & 4177 & 8 & (9.3. 90.7) & 9.7 \\
 \bottomrule
\end{tabular}}
\end{table}

\begin{table}[ht!]
\centering
\caption{Descriptive information for the large datasets ($lds$). This information includes the sample size $m$, the number of variables $n$, the proportion of examples from the minority and majority class \%class(min,maj), and the imbalance ratio $IR$.} \label{tab3}
\resizebox{.99\textwidth}{!}{
\begin{tabular}{llcccc}
\toprule
 code & Dataset & $m$ & $n$ & \%class(min,maj) & $IR$ \\
\midrule
 lds1 & poker 0 vs 5 & 412,600 & 10 & (0.4. 99.6) & 250.6 \\
 lds2 & poker 0 vs 2 & 450,022 & 10 & (8.7. 91.3) & 10.5 \\
 lds3 & covtype7 & 464,677 & 54 & (3.5. 96.5) & 27.6 \\
 lds4 & susy ir16 & 2,881,684 & 18 & (6.0, 94.0) & 16 \\
 lds5 & Hepmass ir16 & 5,578,255 & 27 & (6.0, 94.0) & 16 \\
 \bottomrule
\end{tabular}}
\end{table}

For the first set of experiments, the goal was to assess the predictive performance of the classifiers with the preprocessed data generated by SMOTENN on small and medium-size datasets. To this end, we considered alternative resampling techniques that are not designed for imbalanced Big Data classification. The proposed SMOTENN method was implemented using the exact computation of the neighborhood based on the Euclidean distance and without considering the MapReduce framework. The following alternative methods were considered: SMOTE \cite{Cha2}, Borderline-SMOTE (B-SMO) \cite{Han2005}, Safe-level SMOTE (SL-SMO)\cite{Bunkhumpornpat2009}, ADASYN \cite{He2008}, Adaptive Neighbor SMOTE (AN-SMO) \cite{ANSMOTE17}, Density Based SMOTE (D-SMO)\cite{bun11}, MWMOTE \cite{Barua2014}, and Relocating Safe-level SMOTE (RSL-SMO) \cite{RSLSMOTE16}. For these SMOTE variants, we set $k=5$ as the default value for the number of neighbors, as suggested in \cite{MALDONADO2022108511}. 

The following classifiers were considered: $k$-nearest neighbors ($k$-NN) with $k=5$, linear support vector machine (SVM) with $C=1$, and logistic regression (logit). A detailed description of these classifiers can be found in \cite{Hastie2009}. The default parameters have been considered in previous studies for class-imbalance classification \cite{LaLoVa2019SMOTE,MALDONADO2022108511}.

The geometric mean (g-mean) is used as the main performance metric, computed as $\sqrt{sensitivity \cdot specificity}$, with sensitivity and specificity being the true positive rate and true negative rate, respectively. This measure has been widely used in binary classification in the context of class-imbalanced classification \cite{Kim2016}. 

For the second set of experiments, we implemented the SMOTENN algorithm using the MapReduce framework and the approximate computation of the distances for the definition of the neighborhood. SMOTENN is compared with RUS, ENN, Approx-SMOTE (SMOTE), the combination of RUS and Approx-SMOTE (RUS+SMO), and the combination of ENN and Approx-SMOTE (ENN+SMO). The implementation of ENN was also made using the approximated distance function proposed in \cite{Maillo2020}. The classifiers considered in this study were the decision trees (DT) and random forest (RF) available in Spark's MLlib library \cite{Mllib2016}. 

For both the ENN and SMOTE methods, we set $K=5$ as the default. The RUS strategy was implemented in such a way that the IR in the preprocessed training set was 1. Regarding our approach, we explored different parameter configurations in the spirit of defining their influence and obtaining a recommended default configuration and/or a limited range of values to explore. The following parameter values were explored for the SMOTENN method: the number of nearest neighbors $K=\{5,11,15\}$, the amount of oversampling for the SMOTE step $N=\{1,3\}$, and the proportion of the majority class after the RUS step $p=\{3,4,5,6\}$. As an illustrative example, $p=3$ corresponds to having a 3:1 proportion of majority and minority class examples. 

Regarding the performance metric, we consider the $g$-mean, which can be computed as the geometric mean of the precision and the recall. This metric provides a suitable balance for the correct classification of the two classes, which is important in class-imbalanced classification \cite{LUQUE2019216}. For the small and medium-size datasets, 10-fold cross-validation was conducted using this metric, and the average performance is reported. In contrast, 5-fold cross-validation was performed for the large datasets.

\subsection{Result summary for the small and medium-size datasets}\label{ss_summary1}

For the small and medium-size datasets, the average g-mean is reported for each resampling technique and classifier in Appendix A. To summarize this information, we consolidate the results for the various classifiers in Table \ref{tabHolmSDS} and Table \ref{tabHolmMDS} for the small and medium-size datasets, respectively. Each table presents the following for each resampling technique:
\begin{itemize}
\item Its average rank according to its relative position based on the g-mean measure for all datasets and classifiers. Note that the best resampling method, i.e., the one that achieves the largest g-mean, has a rank of one for a given experiment. Table \ref{tabHolmSDS} encompasses 66 experiments/comparisons (22 datasets and three classifiers), while Table \ref{tabHolmMDS} includes 24 experiments (eight datasets and three classifiers).
\item The average g-mean x100 with its corresponding standard deviation.
\item The $p$-values obtained by the Holm test and the outcome of the test (reject or not). This test was suggested for comparisons between machine learning strategies in combination with the Friedman test with Iman-Davenport correction \cite{MALDONADO2022108511}. The chi-squared statistics computed for the Friedman test were 21.08 and 10.56 for the small and medium-size datasets, respectively, rejecting the null hypothesis of equal ranks for all resampling methods, with p values close to zero ($<0.0001$). This outcome suggests that there are methods able to statistically outperform others, which is confirmed in Tables \ref{tabHolmSDS} and \ref{tabHolmMDS} with the Holm test. This latter test analyzes the pairwise performance between each resampling technique and the one with the best ranking, and its p value is contrasted with a threshold $\alpha/(j-1)$, with significance levels $\alpha=0.05$ and $j=2,\ldots,9$. 
\item The number of wins, ties, or losses (W/T/L) in relation to the one with the best rank in terms of g-mean (the proposed SMOTENN for both groups of datasets).
\end{itemize} 

\begin{table}[ht!]
\centering
\caption{Results of Holm's post-hoc test for pairwise comparisons. Small datasets.} \label{tabHolmSDS}
\resizebox{.99\textwidth}{!}{
\begin{tabular}{lcccccc}
\toprule
 Method & Rank & g-mean & Holm's p & $\alpha/(j-1)$ & outcome & W/T/L \\
\midrule
 SMOTENN & 1.77 & 86.29 $\pm$ 11.42 & - & - & - & -\\
 AN-SMO & 4.86 & 77.43 $\pm$ 22.59 & $<$0.001 & 0.050 & reject & 7/1/58 \\
 SMOTE & 4.89 & 77.81 $\pm$ 23.99 & $<$0.001 & 0.025 & reject & 8/1/57 \\
 MWMOTE & 5.08 & 77.71 $\pm$ 23.13 & $<$0.001 & 0.017 & reject & 7/0/59 \\
 B-SMO & 5.17 & 77.78 $\pm$ 23.77 & $<$0.001 & 0.013 & reject & 7/0/59 \\
 SL-SMO & 5.34 & 76.70 $\pm$ 24.43 & $<$0.001 & 0.010 & reject & 4/0/62 \\
 RSL-SMO & 5.66 & 75.66 $\pm$ 24.06 & $<$0.001 & 0.008 & reject & 5/0/61 \\
 D-SMO & 5.95 & 76.13 $\pm$ 23.11 & $<$0.001 & 0.007 & reject & 5/0/61 \\
 ADASYN & 6.29 & 75.63 $\pm$ 23.18 & $<$0.001 & 0.006 & reject & 7/0/59 \\
 \bottomrule
\end{tabular}}
\end{table}

\begin{table}[ht!]
\centering
\caption{Results of Holm's post-hoc test for pairwise comparisons. Medium-size datasets.} \label{tabHolmMDS}
\resizebox{.99\textwidth}{!}{
\begin{tabular}{lcccccc}
\toprule
 Method & Rank & g-mean & Holm's p & $\alpha/(j-1)$ & outcome & W/T/L \\
\midrule
 SMOTENN & 1.54 & 88.79 $\pm$ 9.92 & - & - & - & -\\
 SMOTE & 4.44 & 80.54 $\pm$ 19.89 & $<$0.001 & 0.050 & reject & 2/1/21 \\
 MWMOTE & 4.94 & 78.45 $\pm$ 22.19 & $<$0.001 & 0.025 & reject & 0/2/22 \\
 SL-SMO & 5.15 & 80.51 $\pm$ 18.41 & $<$0.001 & 0.017 & reject & 1/2/21 \\
 ADASYN & 5.19 & 77.55 $\pm$ 22.50 & $<$0.001 & 0.013 & reject & 1/2/21 \\
 AN-SMO & 5.42 & 78.44 $\pm$ 21.56 & $<$0.001 & 0.010 & reject & 1/1/22 \\
 B-SMO & 5.79 & 77.83 $\pm$ 24.78 & $<$0.001 & 0.008 & reject & 1/2/21 \\
 RSL-SMO & 5.88 & 76.88 $\pm$ 24.37 & $<$0.001 & 0.007 & reject & 0/2/22 \\
 D-SMO & 6.67 & 76.28 $\pm$ 24.65 & $<$0.001 & 0.006 & reject & 0/2/22 \\
 \bottomrule
\end{tabular}}
\end{table}

In Tables \ref{tabHolmSDS} and \ref{tabHolmMDS}, we clearly observe that the proposed SMOTENN consistently delivers superior performance in terms of g-mean, statistically outperforming other oversampling techniques. In terms of average rank (second column), the method posts scores of 1.77 and 1.54 for the low and medium-size datasets, respectively. For context, the next best methods are Adaptive Neighbor SMOTE (AN-SMO, average rank 4.86) and traditional SMOTE (average rank 4.44) for the low and medium-size datasets, respectively. Additionally, SMOTENN boasts an average g-mean (third column) surpassing the second-ranked method by 8.25 and 8.86 for the low and medium-size datasets, respectively. The proposed approach only falls short against the second-ranked method in seven out of 66 cases and in two out of 24 comparisons for the low and medium-size datasets, respectively.

\subsection{Result summary for the large datasets}\label{ss_summary2}

For the large datasets, we replicated the analysis considering resampling techniques and classifiers from the MLlib library \cite{Mllib2016}. Table \ref{tabHolmLDS} summarizes the results for the six resampling techniques and two baseline classifiers (decision trees and random forest). 

\begin{table}[ht!]
\centering
\caption{Results of Holm's post-hoc test for pairwise comparisons. Large datasets.} \label{tabHolmLDS}
\resizebox{.99\textwidth}{!}{
\begin{tabular}{lcccccc}
\toprule
 Method & Rank & g-mean & Holm's p & $\alpha/(j-1)$ & outcome & W/T/L \\
\midrule
SMOTE & 2.10 & 80.54 $\pm$ 22.29 & - & - & - & - \\
SMOTENN & 2.75 & 78.32 $\pm$ 24.54 & 0.437 & 0.050 & not reject & 4/0/6 \\
RUS & 3.00 & 78.74 $\pm$ 23.27 & 0.282 & 0.025 & not reject & 3/0/7 \\
RUS+SMOTE & 3.10 & 77.13 $\pm$ 21.84 & 0.232 & 0.017 & not reject & 3/0/7 \\
ENN & 4.85 & 60.89 $\pm$ 32.85 & 0.001 & 0.013 & reject & 0/0/10 \\
ENN+SMOTE & 5.20 & 69.10 $\pm$ 30.35 & 0.000 & 0.010 & reject & 1/0/9 \\
\bottomrule
\end{tabular}}
\end{table}

The statistic of the Friedman test with Iman-Davenport correction was 7.053 with a p value below 0.001, concluding that the best method is able to statistically outperform some of the others. For this set of experiments, we observe in Table \ref{tabHolmLDS} that SMOTE achieves the best overall performance, with the proposed SMOTENN method being the second-best strategy with an average rank of 2.75. The third best strategy is random undersampling, confirming our hypothesis that RUS is an excellent alternative for imbalanced Big Data classification. Nevertheless, the first four approaches are very close to each other in terms of average performance.

The detailed results in terms of the g-mean are reported for each resampling technique and classifier in Table \ref{tabDetLDS}. In this table, we observe that no clear winner can be depicted from these experiments. However, the proposed SMOTENN achieves a competitive predictive performance, superior to most resampling methods on average, confirming the positive results achieved in the previous experiments on smaller datasets. 

\begin{table}[ht!]
\centering
\caption{Performance summary for the large datasets. G-mean as performance measure.} \label{tabDetLDS}
\resizebox{.99\textwidth}{!}{
\begin{tabular}{lcccccc}
\toprule
 Dataset & ENN & ENN+SMO & RUS & RUS+SMO & SMOTE & SMOTENN \\
\midrule
 \multicolumn{ 4}{l}{DT as baseline classifier} & & & \\
 lds1 & 96.963 & 87.454 & 93.787 & 93.735 & 99.849 & {\bf 99.855} \\
 lds2 & 12.155 & 93.369 & 95.445 & {\bf 96.229} & 93.368 & 93.369 \\
 lds3 & 49.547 & 45.662 & 60.100 & 61.813 & {\bf 68.423} & 58.291 \\
 lds4 & 28.248 & 25.371 & 44.219 & 44.461 & {\bf 47.090} & 46.909 \\
 lds5 & 63.818 & 88.286 & 97.218 & 88.319 & {\bf 98.148} & 96.930 \\
 \multicolumn{ 4}{l}{RF as baseline classifier} & & & \\
 lds1 & 94.206 & 91.725 & 98.380 & 99.698 & 99.702 & {\bf 99.899} \\
 lds2 & 93.368 & 93.368 & {\bf 95.902} & 95.735 & 93.369 & 95.566 \\
 lds3 & 46.757 & 44.010 & 63.648 & {\bf 65.452} & 60.640 & 50.017 \\
 lds4 & 26.775 & 24.310 & 42.853 & 42.530 & {\bf 46.926} & 45.390 \\
 lds5 & 97.039 & 97.398 & 95.798 & 83.287 & {\bf 97.916} & 96.981 \\
 \bottomrule
\end{tabular}}
\end{table}

\subsection{Sensitivity analysis and running times}\label{ss_sensitivity}

The parameter $K$ is arguably the main tuning feature of the proposed SMOTENN method. It represents the neighborhood in which samples are removed (ENN step) or created via interpolation (SMOTE step), and it is not necessarily similar to the neighborhood defined by SMOTE. Table \ref{tabSensK} presents the results for the various $K$ values explored for the large datasets. 

\begin{table}[ht!]
\centering
\caption{Sensitivity analysis for parameter $K$ for the five large datasets and the two classifiers. Results in terms of g-mean for the large datasets.} \label{tabSensK}
\begin{tabular}{ccccccc}
\toprule
Classifier & $k$ & lds1 & lds2 & lds3 & lds4 & lds5 \\
\midrule
\multicolumn{ 1}{c}{DT} & 5 & 99.852 & 93.368 & 56.883 & 46.909 & 96.930 \\
\multicolumn{ 1}{c}{} & 11 & 99.849 & 93.369 & 56.496 & 45.573 & 96.842 \\
\multicolumn{ 1}{c}{} & 15 & 99.845 & 93.369 & 58.291 & 45.902 & 96.857 \\
\multicolumn{ 1}{c}{RF} & 5 & 99.899 & 95.563 & 41.890 & 45.390 & 96.955 \\
\multicolumn{ 1}{c}{} & 11 & 99.899 & 95.566 & 49.977 & 43.805 & 96.939 \\
\multicolumn{ 1}{c}{} & 15 & 98.839 & 95.563 & 50.017 & 43.861 & 96.981 \\
\bottomrule
\end{tabular}
\end{table}

In Table \ref{tabSensK}, we observe very stable results in terms of g-mean for the various $K$ values. We suggest $K=5$ as a default configuration for the SMOTENN method to alleviate the need to find optimal parameters. This configuration achieved the best overall performance.

Finally, we analyze the complexity and running times for all resampling methods on the large datasets. The complexity of finding the $k$-NN for a given sample is $\mathcal{O}(n \cdot m)$, with $n$ being the sample size and $m$ the number of variables. This is because the method computes the distance between the query sample and every other example in the training set \cite{Maillo2020}. For the SMOTE algorithm, the complexity becomes $\mathcal{O}(n\log_2 n)$, with $n$ being the cardinality of $\mathcal{S}_+$ \cite{xiaolong2019over}. Our method has similar complexity, although it searches a neighborhood in $\mathcal{S}_-^* \cup \mathcal{S}_+$ ($n=|\mathcal{S}_-^* \cup \mathcal{S}_+|$). The approximate $k$-NN search made by the hybrid spill tree can speed up the process as the tree has a depth of $\mathcal{O}(\log_2 n)$ \cite{Maillo2020}. Regarding memory consumption, the implementation of the algorithm requires both the training and test sets to be stored in the main memory, which can increased easily in case of large datasets.

The analysis of the running times was performed on an Amazon Web Services (AWS) Elastic MapReduce (EMR) cluster, consisting of a master instance and ten worker instances. The master node was a m5.xlarge instance (4 vcpu, 16 GB memory), while the worker nodes were c5.5xlarge instances (16 vcpu, 32 GB memory) each. Training times for the classifiers are not considered, as they are very similar for all resampled subsets. The codes were implemented using Scala 2.11 and Spark 2.4.8 on the Amazon Linux operating system (64 bits). The results are presented in Table \ref{tabTimes}.

\begin{table}[ht!]
\centering
\caption{Running times, in seconds, for all resampling strategies. Large datasets.} \label{tabTimes}
\begin{tabular}{lccccc}
\toprule
 Method & lds1 & lds2 & lds3 & lds4 & lds5 \\
\midrule
 ENN & 19''.2 & 19''.8 & 16''.5 & 68''.2 & 339''.3 \\
 ENN+SMO & 26''.8 & 33''.0 & 27''.2 & 96''.3 & 757''.1 \\
 RUS & 1''.4 & 1''.3 & 1''.4 & 3''.6 & 3''.2 \\
 RUS+SMO & 4''.5 & 8''.6 & 6''.0 & 18''.2 & 52''.9 \\
 SMOTE & 3''.8 & 6''.4 & 5''.0 & 12''.2 & 33''.8 \\
 SMOTENN & 5''.5 & 11''.6 & 8''.7 & 32''.3 & 551''.8 \\
 \bottomrule
\end{tabular} 
\end{table}

In Table \ref{tabTimes}, we observe negligible running times in general, which are less than two minutes for all datasets and methods, with the exception of lds5. The fastest times are for lds1, lds2, and lds3, which have roughly the same sample size (around 500,000 examples; best-case scenario). Their times range from five to eleven seconds, varying by the number of variables and imbalance ratio. Comparing lds4 and lds5, even though lds5 is only twice the size of lds4, the running time for SMOTENN is 17 times longer (worst-case scenario). This indicates that running times do not increase linearly with sample size.

The proposed approach has relatively similar running times in comparison to Approx-SMOTE, being faster than ENN and SMOTE performed independently and faster than ENN in most cases. The running times for SMOTENN average 38.8\% of ENN and SMOTE when performed independently, with best and worst cases being 20.4\% and 72.9\%, respectively. This result confirms the virtues of SMOTENN in terms of computational efficiency and performance.

\section{Conclusions}\label{sec:conc}

In this paper, a novel strategy for data resampling is presented for dealing with the class-imbalance problem. The proposed SMOTENN method performs intelligent undersampling and oversampling in the same pass over the data. After a neighborhood of size $k$ is defined, samples from the majority class are removed in case they are outnumbered by the examples from the minority class in the neighborhood (ENN step). Meanwhile, synthetic elements are created for the minority class via interpolation (SMOTE step). The most time-consuming stage of the method is the computation of the neighborhood, which is performed via a scalable distance function on MapReduce to confer efficiency to the method.

Our results show the benefits of SMOTENN in terms of scalability and effectiveness on 35 datasets of different sizes. This method performs better in terms of the g-mean compared to well-known resampling techniques. Its advantage relies on its ability to reduce noise in borderline regions efficiently. Robust results are observed in terms of the parameters, showing that a default configuration of the method can achieve the best results on average. 

Although our method is able to perform hybrid resampling in minutes, a simple method such as RUS performs very well on large-scale settings. Random undersampling can perform the resampling task in less than four seconds for all datasets, suggesting that the creation of synthetic instances of the minority class is not strictly necessary when millions of samples are available. However, RUS has some limitations when dealing with noise, and intelligent undersampling strategies can be useful in Big Data environments. We consider the development of novel intelligent undersampling methods to be a very interesting avenue for future research.

The proposed approach can be extended to a multiclass scenario where the target variable has more than two levels \cite{SLEEMANIV2021106598}. Within this context, one could determine one or multiple underrepresented classes as candidates for the oversampling phase. We propose it as a possible avenue for future research endeavors.

A limitation of the proposed approach and the state-of-the-art imbalanced Big Data classification is that it does not consider high-dimensional datasets. The inclusion of several hundreds of potentially irrelevant/redundant variables can negatively affect the performance of resampling strategies \cite{MALDONADO2022108511}. SMOTENN can be extended to address this issue by adapting techniques that handle class-imbalanced high-dimensional settings. For example, FW-SMOTE \cite{MALDONADO2022108511} weighs the features according to their relevance in the definition of the neighborhood defined for interpolation in SMOTE. This approach can be extended easily to Big Data by using an approximate weighted distance function and a fast feature ranking method. Such a study represents a valuable opportunity for future work.

\bibliographystyle{IEEEtran}
\bibliography{biblio-BD-IMB}

\clearpage
\appendix
\renewcommand{\theequation}{\thesection.\arabic{equation}} 
\setcounter{equation}{0} 
\renewcommand{\thetable}{\thesection.\arabic{table}} 
\setcounter{table}{0}

\section{Detailed results for the small and medium-size datasets}\label{sec:detaileda}

Tables \ref{tabDetSDS1} to \ref{tabDetSDS3} present the results for the small datasets using $k$-NN, SVM, and LR as classification models. Each table presents the average g-mean considering 10-fold crossvalidation obtained for each resampling method. The best performance is emphasized in bold type. 

\begin{table}[ht!]
\centering
\caption{Performance summary using $k$-nearest neighbors as the baseline classifier. Small datasets. G-mean as the performance measure.} \label{tabDetSDS1}
\resizebox{.99\textwidth}{!}{
\begin{tabular}{lccccccccc}
\toprule
 Dataset & SMOTE & B-SMO & SL-SMO & ADASYN & AN-SMO & D-SMO & MWMOTE & RSL-SMO & SMOTENN \\
\midrule
 sds1 & 87.5 & 87.5 & 98.6 & 98.6 & 98.6 & 87.5 & 98.6 & {\bf 98.6} & 94.9 \\
 sds2 & 0.0 & 0.0 & 0.0 & 77.8 & 77.8 & {\bf 87.7} & {\bf 87.7} & 77.2 & 72.2 \\
 sds3 & {\bf 68.4} & 65.9 & 62.5 & 6.7 & 9.4 & 8.8 & 9.4 & 0.0 & 63.6 \\
 sds4 & 90.9 & 90.1 & 90.6 & 88.0 & 90.9 & 90.9 & 90.4 & 90.9 & {\bf 91.0} \\
 sds5 & 90.7 & 90.0 & 90.7 & 90.0 & 90.5 & 83.7 & 90.7 & 90.7 & {\bf 91.9} \\
 sds6 & 90.9 & 90.7 & 90.9 & 90.6 & 90.9 & 88.0 & 90.5 & 90.7 & {\bf 92.1} \\
 sds7 & 90.9 & 90.3 & 90.9 & 90.9 & 90.8 & 90.9 & 90.2 & 90.9 & {\bf 92.4} \\
 sds8 & 94.7 & 94.9 & 68.5 & 78.7 & 68.7 & 75.5 & {\bf 95.2} & 68.5 & 90.8 \\
 sds9 & 87.6 & 87.4 & 87.8 & 80.1 & 90.0 & 79.8 & 87.0 & 87.8 & {\bf 90.3} \\
 sds10 & 78.9 & 81.1 & 78.4 & {\bf 87.7} & 80.8 & 78.9 & 80.0 & 79.1 & 81.2 \\
 sds11 & 80.1 & 80.1 & 78.7 & {\bf 88.9} & 79.1 & 78.3 & 78.5 & 78.9 & 79.7 \\
 sds12 & 90.7 & 90.4 & 90.8 & 87.9 & 91.1 & 90.9 & 90.4 & 91.1 & {\bf 93.3} \\
 sds13 & {\bf 91.2} & 89.3 & 89.7 & 86.0 & 89.7 & 87.3 & 89.7 & 89.7 & 90.6 \\
 sds14 & {\bf 91.0} & {\bf 91.0} & 90.8 & 88.1 & {\bf 91.0} & 90.7 & 90.8 & {\bf 91.0} & 90.5 \\
 sds15 & 88.0 & 87.6 & 87.5 & 86.1 & 79.7 & 79.9 & 88.0 & 77.9 & {\bf 90.0} \\
 sds16 & 83.5 & 85.3 & 84.1 & 86.5 & 83.7 & 84.8 & 84.9 & 84.2 & {\bf 90.2} \\
 sds17 & 79.8 & 79.8 & 86.8 & 79.7 & 86.8 & 86.6 & 79.7 & 79.7 & {\bf 97.9} \\
 sds18 & 51.3 & 48.6 & 33.2 & 25.2 & 27.1 & 25.2 & 37.8 & 23.0 & {\bf 65.4} \\
 sds19 & 92.2 & 94.3 & 92.3 & 92.6 & 93.5 & 93.6 & 91.9 & 94.5 & {\bf 95.3} \\
 sds20 & 88.7 & 89.1 & 85.9 & 87.4 & 87.8 & 83.7 & 88.7 & 84.2 & {\bf 90.7} \\
 sds21 & {\bf 82.6} & 78.1 & 77.2 & 80.3 & 80.1 & 80.8 & 81.3 & 76.7 & 80.1 \\
 sds22 & 49.0 & 38.9 & 19.1 & 39.5 & 46.6 & 40.7 & 44.3 & 40.9 & {\bf 52.1} \\
\bottomrule
\end{tabular}}
\end{table}

\begin{table}[ht!]
\centering
\caption{Performance summary using support vector machine as the baseline classifier. Small datasets. G-mean as the performance measure.} \label{tabDetSDS2}
\resizebox{.99\textwidth}{!}{
\begin{tabular}{lccccccccc}
\toprule
 Dataset & SMOTE & B-SMO & SL-SMO & ADASYN & AN-SMO & D-SMO & MWMOTE & RSL-SMO & SMOTENN \\
\midrule
 sds1 & 88.9 & 88.2 & 88.2 & 88.2 & 88.2 & 88.9 & 88.9 & 88.2 & {\bf 95.6} \\
 sds2 & 88.2 & 88.9 & {\bf 99.4} & 88.3 & 88.3 & 99.4 & 99.4 & 98.9 & 81.5 \\
 sds3 & 0.0 & 0.0 & 0.0 & 0.0 & 0.0 & 0.0 & 0.0 & 6.9 & {\bf 53.1} \\
 sds4 & 86.8 & 87.1 & 90.5 & 83.9 & 87.0 & 90.2 & 87.1 & 87.5 & {\bf 92.4} \\
 sds5 & 90.4 & 89.6 & 90.6 & 90.5 & 87.7 & 87.5 & 90.5 & 90.5 & {\bf 92.9} \\
 sds6 & 87.7 & 88.3 & 84.8 & 85.5 & 87.7 & 84.8 & 84.2 & 84.6 & {\bf 92.5} \\
 sds7 & 87.5 & 90.4 & 86.9 & 89.6 & 87.5 & 87.5 & 87.0 & 87.5 & {\bf 92.4} \\
 sds8 & 84.9 & {\bf 94.5} & 75.4 & 62.2 & 94.2 & 85.2 & 84.9 & 85.4 & 91.2 \\
 sds9 & 87.4 & 87.4 & 87.9 & 87.0 & 87.7 & 87.6 & 87.6 & 87.7 & {\bf 90.3} \\
 sds10 & {\bf 81.6} & 80.6 & 78.7 & 81.1 & 81.1 & 78.7 & 80.6 & 78.9 & {\bf 81.6} \\
 sds11 & 73.2 & 72.4 & 71.8 & 72.5 & 73.1 & 71.5 & 71.8 & 71.8 & {\bf 79.8} \\
 sds12 & 86.9 & 89.4 & 77.1 & 76.9 & 89.9 & 77.1 & 86.9 & 77.1 & {\bf 90.9} \\
 sds13 & 87.7 & 89.4 & 89.4 & 89.1 & 89.4 & 85.4 & 89.4 & 85.6 & {\bf 91.5} \\
 sds14 & 87.2 & 88.9 & 77.4 & 84.2 & 84.4 & 84.2 & 84.4 & 84.3 & {\bf 91.8} \\
 sds15 & 88.2 & 88.7 & 88.8 & 87.2 & 83.9 & 84.3 & 72.4 & 83.1 & {\bf 89.7} \\
 sds16 & 87.6 & 88.4 & 88.5 & 88.5 & 87.7 & 89.0 & 89.0 & 88.4 & {\bf 89.9} \\
 sds17 & 86.6 & 93.4 & 86.6 & 86.4 & 86.7 & 86.6 & 93.7 & 86.6 & {\bf 98.2} \\
 sds18 & 0.0 & 5.8 & 5.8 & 11.5 & 23.0 & 11.5 & 11.5 & 17.3 & {\bf 56.1}\\
 sds19 & 80.8 & 86.8 & 86.3 & 76.5 & 68.2 & 63.0 & 69.9 & 58.7 & {\bf 94.6} \\
 sds20 & 88.0 & 78.0 & 87.1 & 83.1 & 88.1 & 88.0 & 87.5 & 88.1 & {\bf 88.8} \\
 sds21 & 77.4 & 78.5 & 77.6 & 75.8 & {\bf 80.1} & 79.0 & 79.5 & 79.3 & 79.6 \\
 sds22 & 0.0 & 0.0 & 0.0 & 36.4 & 42.5 & 34.8 & 36.4 & 42.5 & {\bf 51.5} \\
\bottomrule
\end{tabular}}
\end{table}

\begin{table}[ht!]
\centering
\caption{Performance summary using logistic regression as the baseline classifier. Small datasets. G-mean as the performance measure.} \label{tabDetSDS3}
\resizebox{.99\textwidth}{!}{
\begin{tabular}{lccccccccc}
\toprule
 Dataset & SMOTE & B-SMO & SL-SMO & ADASYN & AN-SMO & D-SMO & MWMOTE & RSL-SMO & SMOTENN \\
\midrule
 sds1 & 86.0 & 86.0 & 86.0 & 86.0 & 86.0 & 86.0 & 86.0 & 86.0 & {\bf 95.6} \\
 sds2 & {\bf 97.0} & {\bf 97.0} & {\bf 97.0} & {\bf 97.0} & {\bf 97.0} & {\bf 97.0} & {\bf 97.0} & {\bf 97.0} & 94.7 \\
 sds3 & 50.5 & 57.0 & 49.5 & 16.5 & 13.7 & 16.5 & 16.5 & 7.1 & {\bf 81.9} \\
 sds4 & 86.7 & 88.0 & 86.7 & 80.5 & 86.7 & 86.7 & 89.0 & 86.7 & {\bf 92.2} \\
 sds5 & 89.0 & 86.1 & 86.7 & 86.7 & 83.7 & 86.7 & 86.5 & 86.7 & {\bf 92.8} \\
 sds6 & 90.1 & 88.8 & 87.5 & 86.2 & 87.5 & 87.5 & 87.5 & 87.5 & {\bf 91.7} \\
 sds7 & 86.7 & 89.4 & 87.0 & 90.1 & 87.0 & 87.0 & 87.0 & 87.0 & {\bf 92.7} \\
 sds8 & 69.5 & 70.2 & 79.3 & 59.7 & 79.7 & 72.3 & 82.5 & 60.2 & {\bf 92.9} \\
 sds9 & 87.2 & 87.2 & 87.5 & 87.1 & 87.1 & 87.4 & 87.4 & 87.5 & {\bf 90.1} \\
 sds10 & 83.5 & 82.0 & 83.5 & 82.9 & 83.2 & 83.2 & 82.7 & 83.5 & {\bf 87.0} \\
 sds11 & 80.5 & 80.3 & 80.7 & 79.3 & 80.3 & 80.0 & 80.7 & 80.5 & {\bf 87.8} \\
 sds12 & 90.2 & 90.0 & 87.5 & 80.2 & 90.4 & 87.5 & 90.0 & 87.2 & {\bf 93.9} \\
 sds13 & 89.4 & 86.7 & 89.5 & 87.4 & 89.7 & 89.4 & 89.4 & 89.7 & {\bf 91.8} \\
 sds14 & 87.0 & 88.7 & 84.3 & 84.2 & 84.4 & 84.3 & 87.2 & 84.3 & {\bf 91.6} \\
 sds15 & 88.9 & 88.1 & 88.9 & 85.7 & 77.1 & 77.5 & 79.5 & 75.6 & {\bf 89.6} \\
 sds16 & 86.6 & 87.3 & 87.0 & 87.8 & 84.3 & 88.9 & 88.6 & 85.6 & {\bf 90.7} \\
 sds17 & 86.2 & 85.8 & 86.2 & 86.2 & 86.2 & 86.2 & 86.2 & 86.2 & {\bf 95.6} \\
 sds18 & 69.7 & 66.3 & 60.4 & 15.5 & 17.8 & 15.5 & 13.8 & 8.1 & {\bf 73.6 }\\
 sds19 & 89.9 & 73.7 & 90.9 & 71.7 & 93.2 & 87.3 & 81.2 & 90.7 & {\bf 95.6} \\
 sds20 & 84.4 & 81.4 & 82.6 & 83.9 & 85.9 & 81.7 & 86.6 & 83.9 & {\bf 89.4} \\
 sds21 & 75.4 & 77.0 & 77.1 & 68.9 & 76.6 & 73.7 & 72.9 & 71.1 & {\bf 79.5} \\
 sds22 & 5.7 & 11.4 & 22.7 & 46.6 & 41.8 & 43.9 & 42.2 & 36.1 & {\bf 62.2} \\
\bottomrule
\end{tabular}}
\end{table}

Finally, Table \ref{tabDetMDS} presents the results for the medium-size datasets in terms of g-mean. The best performance is also highlighted in bold. 

\begin{table}[ht!]
\centering
\caption{Performance summary for the medium-size datasets. G-mean as the performance measure.} \label{tabDetMDS}
\resizebox{.99\textwidth}{!}{
\begin{tabular}{lccccccccc}
\toprule
 Dataset & SMOTE & B-SMO & SL-SMO & ADASYN & AN-SMO & D-SMO & MWMOTE & RSL-SMO & SMOTENN \\
\midrule
 \multicolumn{ 4}{l}{kNN as baseline classifier} & & & & & & \\
 mds1 & 88.9 & 88.2 & 89.4 & 89.0 & 88.5 & 89.2 & 88.8 & 89.5 & {\bf 91.5} \\
 mds2 & 51.4 & 38.5 & 38.2 & 50.5 & 45.3 & 36.8 & 44.9 & 32.8 & {\bf 73.0} \\
 mds3 & 88.6 & 87.4 & 88.2 & 88.5 & 88.3 & 86.5 & 88.7 & 87.4 & {\bf 90.4} \\
 mds4 & 70.9 & 66.9 & 66.7 & 21.9 & 27.6 & 18.9 & 28.7 & 17.8 & {\bf 83.9} \\
 mds5 & 94.4 & 94.6 & 91.4 & 86.5 & 91.4 & 91.7 & 95.6 & 91.5 & {\bf 96.9} \\
 mds6 & 99.6 & 99.6 & 99.6 & 99.6 & 99.6 & 99.6 & 99.6 & 99.6 & {\bf 100} \\
 mds7 & {\bf 99.4} & 99.3 & {\bf 99.4} & 99.3 & {\bf 99.4} & 98.8 & 99.2 & 99.2 & 99.3 \\
 mds8 & 68.3 & 67.0 & 64.5 & 68.4 & 70.9 & 66.6 & 69.3 & 65.3 & {\bf 77.1} \\
 \multicolumn{ 4}{l}{SVM as baseline classifier} & & & & & & \\
 mds1 & 89.6 & 88.2 & 89.8 & 89.3 & 89.7 & 88.6 & 88.5 & 89.9 & {\bf 90.8} \\
 mds2 & 46.1 & 42.3 & 46.1 & 44.4 & 44.3 & 39.3 & 46.1 & 45.7 & {\bf 71.7} \\
 mds3 & 88.9 & 88.3 & 88.6 & 89.3 & 88.2 & 89.3 & 88.9 & 88.2 & {\bf 91.5} \\
 mds4 & 42.1 & 4.5 & 57.6 & 49.1 & 49.3 & 46.6 & 50.6 & 52.6 & {\bf 84.4} \\
 mds5 & 92.7 & 89.7 & 90.0 & 85.3 & 87.2 & 91.2 & 92.7 & 91.3 & {\bf 96.8}\\
 mds6 & {\bf 100} & {\bf 100} & {\bf 100} & {\bf 100} & {\bf 100} & {\bf 100} & {\bf 100} & {\bf 100} & {\bf 100} \\
 mds7 & 99.1 & 99.4 & 99.2 & {\bf 99.7} & 99.1 & 99.2 & 99.1 & 99.2 & 99.2 \\
 mds8 & 73.9 & 67.2 & 73.9 & 75.2 & 73.6 & 73.3 & 72.5 & 73.0 & {\bf 75.9} \\
 \multicolumn{ 4}{l}{Logit as baseline classifier} & & & & & & \\
 mds1 & {\bf 90.6} & 87.7 & 89.8 & 88.6 & 89.7 & 87.7 & 88.9 & 89.9 & 90.5 \\
 mds2 & 34.2 & 41.1 & 52.1 & 41.3 & 53.3 & 40.6 & 43.1 & 46.0 & {\bf 71.0} \\
 mds3 & 88.2 & 89.0 & 87.9 & 89.0 & 88.2 & 88.0 & 88.2 & 87.5 & {\bf 90.2} \\
 mds4 & 68.7 & 65.7 & 68.9 & 52.5 & 52.6 & 46.4 & 52.6 & 43.3 & {\bf 83.5} \\
 mds5 & 85.8 & 91.5 & 80.4 & 82.5 & 84.8 & 83.6 & 85.8 & 84.8 & {\bf 96.1} \\
 mds6 & 99.6 & 99.6 & 99.6 & 99.6 & 99.6 & 99.6 & 99.6 & 99.6 & {\bf 100} \\
 mds7 & 98.8 & 99.1 & 98.8 & 99.2 & 99.0 & 99.1 & 99.3 & 99.2 & {\bf 99.3} \\
 mds8 & 73.2 & 73.0 & 72.1 & 72.5 & 72.9 & 70.0 & 72.1 & 71.7 & {\bf 78.0} \\
\bottomrule
\end{tabular}}
\end{table}

\end{document}